%% file: main.tex
\documentclass[10pt,twocolumn,letterpaper]{article}

\usepackage[pagenumbers]{cvpr} 

\usepackage{graphicx}
\usepackage{amsmath}
\usepackage{amssymb}
\usepackage{booktabs}
\usepackage{algorithm}
\usepackage{algorithmic}
\usepackage{tabularx}
\usepackage[numbers,sort&compress]{natbib}

\usepackage[pagebackref,breaklinks,colorlinks,citecolor=cvprblue]{hyperref}

\usepackage[capitalize]{cleveref}
\crefname{section}{Sec.}{Secs.}
\Crefname{section}{Section}{Sections}
\Crefname{table}{Table}{Tables}
\crefname{table}{Tab.}{Tabs.}

\input{preamble}

\def\paperID{4058} 
\def\confName{CVPR}
\def\confYear{2024}

\begin{document}

\title{You Only Explain Once}

\author{David A. Kelly, Hana Chockler, Nathan Blake, Aditi Ramaswamy,\\ Melane Navaratnarajah, Aaditya Shivakumar\\
Department of Informatics\\
King's College London\\
{\tt\small david.a.kelly, hana.chockler, nathan.blake, aditi.ramaswamy,}\\{\tt\small melane.navarathnarajah, aaditya.shivakumar AT kcl.ac.uk}
\and
Daniel Kroening\\
AWS\\
{\tt\small drk@amazon.co.uk}
}

\makeatletter
\let\@oldmaketitle\@maketitle
\renewcommand{\@maketitle}{\@oldmaketitle
\begin{figure}[H]
\setlength{\linewidth}{\textwidth}
\setlength{\hsize}{\textwidth}
\centering
 \begin{subfigure}[t]{0.3\textwidth}
        \centering
        \includegraphics[scale=0.95]{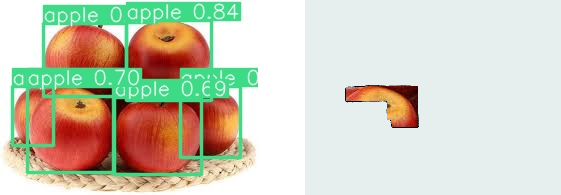}
        \caption{An explanation for an apple}\label{subfig:yoloapple}
    \end{subfigure}%
    ~ 
    \begin{subfigure}[t]{0.3\textwidth}
        \centering
        \includegraphics[scale=1.05]{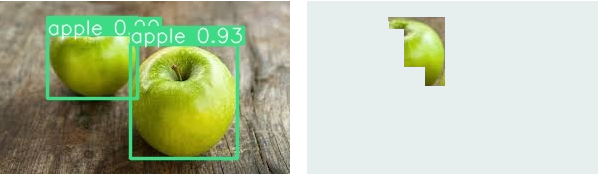}
        \caption{The upper curve of an apple is sufficient}\label{subfig:yorexapple}
    \end{subfigure}
    ~
    \begin{subfigure}[t]{0.3\textwidth}
        \centering
        \includegraphics[scale=0.83]{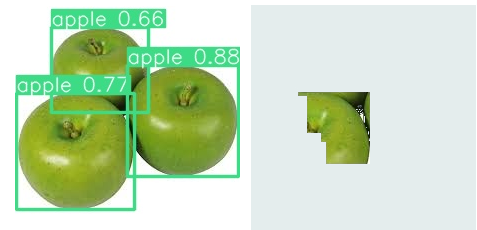}
        \caption{The top of an apple is sufficient}\label{subfig:appleheatmap}
    \end{subfigure}    
    \caption{Three different explanations for an apple, provided by \yorex. While all explanations are strongly rectilinear, they reveal that only a section of an apple is sufficient for classification, regardless of color.}\label{fig:applepx}
\label{fig:accept}%
\end{figure}
\bigskip}%
\makeatother

\maketitle

\begin{figure*}[tp]
\end{figure*}

\maketitle

\begin{abstract}

In this paper, we propose a new black-box explainability algorithm and tool, \yorex, for efficient explanation of the outputs of object detectors. The new algorithm computes explanations for all objects detected in the image simultaneously. Hence, compared to the baseline, the new algorithm reduces the number of queries by a factor of $10\times$ for the case of ten detected objects. The speedup increases further with with the number of objects.
Our experimental results demonstrate that \yorex can explain the outputs of \yolo with a negligible overhead over the running time of \yolo. We also demonstrate similar results for explaining \ssd and \fastrcnn. The speedup is achieved by avoiding backtracking by combining aggressive pruning with a causal analysis.

\end{abstract}

\input{intro}

\input{related}
\input{overview}
\input{algorithm}

\input{results}

\input{conclusions}

\newpage
{\small
\bibliographystyle{ieeenatcvpr}
\bibliography{all}
}

\end{document}

%% file: preamble.tex
\usepackage{tikz}
\usetikzlibrary{calc, fit, positioning, shadows}
\usepackage{dsfont}
\usepackage{pgf}
\usepackage{pgfplots}
\pgfplotsset{width=5cm,compat=1.9}
\pgfplotsset{every tick label/.append style={font=\tiny}}

\definecolor{cvprblue}{rgb}{0.21,0.49,0.74}

\usepackage{pifont}
\Crefname{ALC@unique}{Line}{Lines}

\newcommand{\rex}{\textsc{r}\textnormal{e}\textsc{x}\xspace}
\newcommand{\yorex}{{\textsc{anon}-\rex}\xspace}
\newcommand{\yolo}{\textsc{yolo}\xspace}
\newcommand{\ssd}{\textsc{ssd}\xspace}
\newcommand{\gradcam}{{\sc g}\textnormal{rad}-{\sc cam}\xspace}
\newcommand{\lime}{{\sc lime}\xspace}
\newcommand{\shap}{{\sc shap}\xspace}

\newcommand{\eigencam}{\textsc{e}\textnormal{igen}\textsc{c}\textnormal{am}\xspace}
\newcommand{\eigencamyolo}{\textsc{e}\textnormal{igen}\textsc{c}\textnormal{am}-\yolo}
\newcommand{\deepcover}{\textsc{d}\textnormal{eep}\textsc{c}\textnormal{over}\xspace}
\newcommand{\fastrcnn}{\textsc{f}\textnormal{aster}\,\textsc{r}-\textsc{cnn}\xspace}
\newcommand{\rcnn}{\textsc{r}-\textsc{cnn}\xspace}

\newcommand{\commentout}[1]{}

\newtheorem{observation}{Observation}
\newtheorem{claim}{Claim}
\newtheorem{remark}{Remark}


%% file: intro.tex
\section{Introduction}\label{sec:intro}

Neural networks
are now a primary building block of most computer vision systems. The opacity of
neural networks creates demand for explainability techniques, which attempt to provide insight into
\emph{why} a particular input yields a particular observed output.
Beyond increasing a user's confidence in the output, as well as their trust
in the AI system, these insights help to uncover subtle recognition errors
that are not detectable from the output alone~\cite{CKS21}.

Explainability is especially important for object detection.
Object detectors are fast networks,
localizing and identifying objects within a given image. Object detectors are crucial for 
self-driving vehicles and advanced driver-assistance systems, as well as many other applications, as they are fast enough to
compute results in real time. The state of the art in the field is \yolo~\cite{yolo}, an object detector that localizes and labels objects on an input image within one pass.
Other notable object detectors include \ssd~\cite{ssd}, which is also a one-pass object detector, and a number of
two-passes object detectors, most popular of which are the variants of \rcnn~\cite{girshick2014rich}. These are slower in general, but recognize more classes of objects.
While there is a large body of work on explaining image classifiers, there is no published research on how to explain their results, despite the prevalence of object detection use-cases.

For image classifiers, the standard form for an explanation is a subset of highest-ranked pixels that is sufficient for the original classification outcome -- typically a small number of contiguous areas. Figure~\ref{fig:accept} illustrates how this idea is adapted to the output of object detectors: there must exist a separate explanation for each detected object, contained within the bounding box computed by the object detector.

In this paper, we present \yorex, an explainability algorithm for object detectors such as \yolo. Explaining the outputs of neural networks is a known hard problem, and it is even harder
to explain outputs of object detectors, as they detect multiple objects in a given image and there are strong performance constraints.
We hence introduce aggressive pruning and biased partitioning of the input, as well as the ability to construct explanations of multiple objects in the image simultaneously in order to compute our explanations efficiently.

As there is no existing black-box explainability tool for object detectors, we 
construct a baseline tool for comparison from \rex, which is an explanability tool for image classifiers~\cite{CKS21}. 
Our experimental results show that \yorex outperforms \rex on \yolo by at least an order of magnitude and is much more scalable with respect to the number of objects detected on the image. 
Furthermore, our experimental results demonstrate that \yorex produces significantly better explanations than the only existing white-box explainability tool for object detectors,
\eigencamyolo~\cite{eigencam}.

To demonstrate that our technique is broadly applicable beyond \yolo, we also present experimental results of \yorex on \ssd and \fastrcnn, both showing similar levels of improvement we observe when using \yolo. Full experimental results, datasets, and the code of \yorex are submitted as a part of the supplementary material.

%% file: related.tex
\section{Related Work}\label{sec:relwork}

Existing algorithms for explaining the output of image classifiers can be roughly divided into black-box algorithms, which are agnostic to the structure of the classifier, and white-box ones, which rely on the access to the internals of the classifier. There are clear advantages to the black-box tools, namely being agnostic to the internal structure and complexity of the classifier. Yet, black-box tools are in general much less efficient than white-box ones, as they rely on querying the classifier many times. Thus, they cannot be used as-is to explain the outputs of object detectors, whose main characteristic is efficiency.

There is a large body of work on algorithms for computing an explanation
for a given output of an image classifier. They can be largely grouped
into two categories: propagation and perturbation.  Propagation-based
explanation methods back-propagate a model's decision to the input
layer to determine the weight of each input
feature for the decision~\cite{springenberg2015striving,
sundararajan2017axiomatic, bach2015pixel, shrikumar2017learning,
nam2020relative}. \gradcam only needs
one backward pass and propagates the class-specific gradient into the final convolutional layer of a 
DNN to coarsely highlight important regions of an input image~\cite{CAM}. 

Perturbation-based explanation approaches introduce perturbations to the input space directly
in search for an explanation. \shap (SHapley Additive exPlanations) 
computes Shapley values of different parts of the input and uses them to rank the
features of the input according to their importance~\cite{lundberg2017unified}.
\lime constructs a small neural network to label the original input and its neighborhood of perturbed
images and uses this network to estimate the importance of different parts of the input
~\cite{lime, datta2016algorithmic,
chen2018learning, petsiuk2018rise, fong2019understanding}. 
Finally, \rex (previously called \deepcover) ranks elements of the image according to their importance for the classification
and uses this ranking to greedily construct a small explanation~\cite{sun2020explaining,CKS21,CKK23}. 
The \deepcover ranking procedure in~\cite{sun2020explaining} uses SFL, 
and is replaced in~\cite{CKS21} by the approximate computation of causal responsibility. The latest version, \rex~\cite{CKK23}, computes
multiple explanations for a given input.

None of the tools mentioned above works with \yolo, or any object detector, natively.
There exists a version of \gradcam for \yolo~\cite{gradcamyolo}; unfortunately, however, it is proprietary and not available for experiments. 
An open source, white-box equivalent is
\eigencamyolo~\cite{eigenyolo}, based on \eigencam~\cite{eigencam} and \yolo v8. The approach taken by  \eigencamyolo works for some images, but often it yields explanations that are too large, highlight the wrong sections of the image, or differ significantly from plain \yolo bounding boxes. A major disadvantage of \eigencamyolo is its reliance on heatmaps as its main method of communicating explanations: an image with many identifiable objects, such as the shelf of wine bottles, becomes a red blur when run through \eigencamyolo's explanation algorithm, and the user obtains no further knowledge about the working of the object detector.


%% file: overview.tex
\section{Overview of \rex}\label{sec:rex}
Causal \underline{R}esponsibility \underline{eX}planations (\rex)~\cite{CKK23,CKS21} (formerly \deepcover) is an XAI tool 
based on the solid foundations of causal reasoning~\cite{Hal48}. 
Essentially, \rex constructs an approximation of a causal explanation~\cite{HP05,Hal48} by first ranking the pixels of a given image
according to their importance for the classification, and then constructing an explanation greedily from the saliency landscape of
the pixels. The ranking is an approximate degree of responsibility~\cite{CH04}, 
computed on the coarsely partitioned image, and updated by refining
the areas of the input with high responsibility, until a sufficiently fine partition is reached. 

To compute the degree of responsibility, \rex generates mutants -- images with some of the parts
masked -- and queries the classifier on these mutants. If a non-masked area is sufficient to get the same classification as the original
image, it is a cause for the classification, with the degree of responsibility of each part based on the number of parts in the area.
For example, if a part by itself is sufficient for the classification, the responsibility of each of its pixels is $1$; 
if, however, the smallest set of
parts needed to obtain the same classification is of size $5$, then each pixel in each part of this set has the responsibility $1/5$.

Essentially, the responsibility is in $[0,1]$, and the higher it is, the more important a given pixel is for the classification.
If a part has no influence on the classification, the responsibility of each of its pixel is $0$, and it is eliminated from further analysis.
The process is repeated over a number of different random partitions, and the result is averaged.
The partitions are randomly produced, but the precise nature of the random distribution is flexible. \rex supports a number of distributions natively, including discrete uniform and beta-binomial. We use uniform for all of our experiments.

\commentout{
\rex first partitions an image into four random rectilinear segments. Some of these partitions are then occluded (i.e. all pixels in the segment set to a masking value) and all possible occluded combinations of segments are passed to the model. We call each new occluded image obtained, a mutant. If certain occlusions do not change the model prediction, then those segments are said to be responsible for the prediction. Because various combinations of the segments are being passed into the model, it will sometimes be the case that more than one such segment is responsible for the classification; in this case the responsibility is shared. All pixels in a lone segment would receive a responsibility of $1.0$ and all pixels in all $S$ segments, which together do not change the prediction, receive a responsibility score of $\frac{1}{S}$. These segments are retained and further segmentations performed, repeating the process recursively to obtain responsibility scores for the pixels until a stopping criterion is met. This is repeated over a number of different initial segmentations, and the overall responsibility for each pixel is cumulatively summed. Each pixel is then min-max normalised so that the maximum pixel responsibility is $1$.
}

%% file: algorithm.tex
\section{\yorex Algorithm}\label{sec:algo}
\yorex introduces a number of modifications to the original \rex algorithm, drastically reducing the number of calls to the object detector and building a saliency landscape for multiple objects at the same time. These changes are agnostic to the internals of the object detector, as they change the \rex algorithm only and do not depend on a particular classification or object detection model. There is one caveat to this agnosticism in that different object detector tools return their predictions in different formats. These different formats need to be handled appropriately in order to provide necessary information to \yorex. 

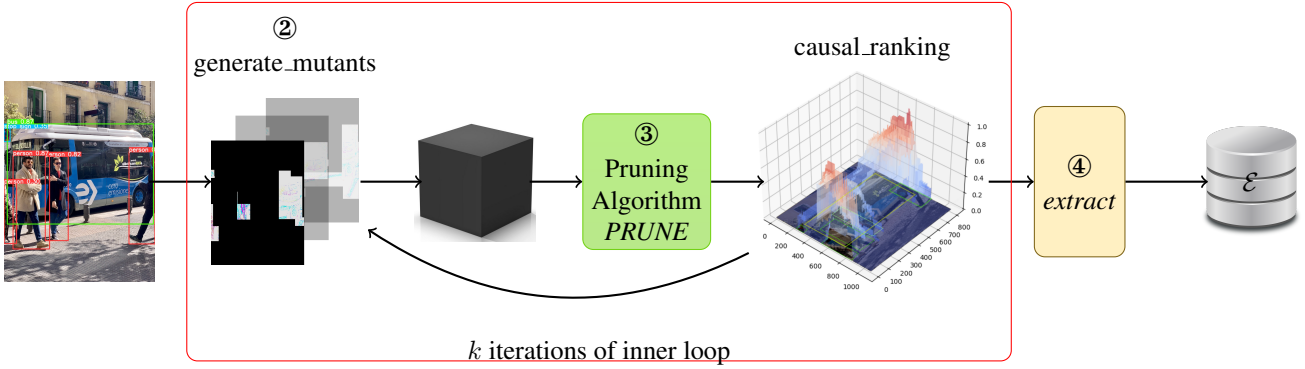
\begin{figure*}
\centering\input{yorex}
\caption{A schematic depiction of our algorithm, returning a set of explanations $\mathcal{E}$, one for each bounding box in the original image (the black
box stands for an object detector). It creates masked mutants for all bounding boxes simultaneously \ding{173}, queries the model, then the \emph{pruning algorithm} \ding{174} selects mutants for further investigation and removes all other mutants. 
\yorex then generates a common saliency landscape of pixels for all objects using the causality-based ranking procedure. 
After $k$ iterations, \ding{175} extracts the explanations from the original bounding boxes and the saliency landscape, 
returning $\mathcal{E}$, the set of explanations for all detected objects in the original image.} 
\label{fig:workflow}
\end{figure*}

We present a high-level view of the algorithm in~\Cref{fig:workflow} and its pseudocode in \Cref{algo:yorex}. Object detection tools return a set of labels and 
bounding boxes for these labels. We use this initial output as a process queue, which we update and refine during the algorithm. 
The algorithm maintains a copy of the original set of predictions and bounding boxes separately from the process queue. This allows us to refer back to the initial bounding boxes when extracting explanations from the saliency landscape. An explanation should not extend beyond the borders of the bounding box of the initial prediction (as we see later in~\Cref{fig:wine}, allowing explanations to bleed out of the original box results in explanations which are difficult to interpret).
The algorithm averages the ranking of pixels over $k$ iterations (for a given parameter $k$). in each iteration calculating the rank of each pixel in each
bounding box, 

\begin{algorithm}[hbt]
    \caption{$\mbox{\yorex}\,(\mathit{x}, \mathcal{N})$}
    \label{algo:yorex}
    \begin{flushleft}
        \textbf{INPUT:}\,\, an image $x$, an object detection model $\mathcal{N}$, number of iterations $k$\\
        \textbf{OUTPUT:}\,\, a set of explanations $\mathcal{E}$ for each bounding box in $\mathcal{N}(x)$
    \end{flushleft}     
        \begin{algorithmic}[1]
         \STATE $\mathcal{D} \leftarrow \mathcal{N}(x)$\label{line:init}
         \STATE $R \leftarrow$ map of $0$'s
        \FOR{$i$ in $0 \ldots k$}
            \STATE $\mathit{Q} \leftarrow \mathcal{D}$
            \STATE initialize all cells of $updated$ to $-1$
            \STATE $\mathcal{E} \leftarrow \emptyset$
            \WHILE{True}
                \STATE $\mathds{P} \leftarrow \mathit{random\_partitions(Q)}$
                \FORALL{$p \in \mathds{P}$}
                    \STATE $\mathit{m} \leftarrow \mathit{generate\_mutant(Q,p)}$
                    \STATE $\mathit{preds} \leftarrow \mathcal{N}(m)$
                    \STATE $(\mathit{updated},\mathit{Q}) \leftarrow \mathit{prune}(preds,\mathcal{D}, updated)$\label{line:update}
                    \IF{($\forall u \in updated: u >= 0$)}
                        \STATE break
                    \ELSIF{($\forall u \in updated: u = -1$)}
                        \STATE break
                    \ENDIF    
                \ENDFOR
            \ENDWHILE
            \STATE $R \leftarrow R + \mathit{causal\_ranking(queue,updated,\mathds{P})}$\label{line:rank}
        \ENDFOR
        \STATE $\mathcal{E} \leftarrow \mathit{extract(R, D)}$
        \RETURN $\mathcal{E}$
        \end{algorithmic}
\end{algorithm}

The \emph{prune} procedure, in combination with the \emph{ranking} procedure, adapted from the one in~\cite{CKS21},  introduces \emph{aggressive pruning}. Essentially, the algorithm first partitions the input image into
$s$ parts and then performs a gradual refinement of the partition based on the \emph{degree of causal responsibility}~\cite{CH04,Hal48} of each part. The original \rex ranking procedure
computes the degree of responsibility of each part and then discards the parts with responsibility $0$. In \yorex, we compute the degree of responsibility greedily, and stop at each level when we find a subset of parts whose union (with the remainder of the image masked) 
elicits the original set of labels from the object detector. All other parts of the bounding box are discarded, and the algorithm progresses
to the next refinement level. 

The original \rex assumes one object per image, whereas \yolo can produce multiple detections. The procedure~\Cref{algo:mutants} mutates all bounding boxes detected in the image at the same time. We add the appropriate set of parts of each bounding box to the mutant. For example, assuming $4$ parts for each bounding box,~\Cref{algo:mutants} creates a mutant where each upper left box for every bounding box is revealed in that mutant. We continue this for all combinations of parts for each bounding box. 

For reasons of efficiency, we group batches of mutants together. To be concrete, we have the first group of combinations as $(0 \ldots s)$, we have each bounding box subdivided into $s$ partitions, numbered appropriately, and we reveal partition $s_i$ for each bounding box to create one mutant. The second group, if required, considers pairs of parts and so on. Grouping in this fashion allows us to balance the benefits of batching (by sending more mutants to the model) without doing too much unnecessary work, \ie considering combinations of partitions when we already have a complete set of passing mutants. 

To keep track of which combination of partitions satisfies the classification,
we maintain an array the same length as $\mathcal{D}$, which holds a combination of parts that is sufficient for a classification ($-1$ means that none were found). As soon as one such combination is found, we mask all other parts and proceed to partition this combination further. 
The following claim is straightforward from the procedure.
\begin{claim}
At any level, the non-masked parts of the image contain explanations for all detected objects in the image.
\end{claim}

\begin{algorithm}[hbt]
    \caption{$\mathit{generate\_mutant(Q, p)}$}
    \begin{flushleft}
        \textbf{INPUT:}\,\, current process queue $Q$, and a set of parts $p$ to be unmasked for each bounding box\\
        \textbf{OUTPUT:}\,\, a mutant $m$
    \end{flushleft}
    \label{algo:mutants}
    \begin{algorithmic}[1]
        \STATE $\mathit{m} \leftarrow \emptyset$
        \FORALL{$q \in \mathit{queue}$}
            \STATE $\mathit{m}\, \leftarrow \mathit{m} \cup$ (unmask $p$ in bounding box of $q$)
            \ENDFOR
        \RETURN $\mathit{m}$
    \end{algorithmic}
\end{algorithm}

There are various subtleties for which we must account. When we mask part of a bounding box and query the model, we do not always get only one prediction in return. For example, if we start with a top-level classification and bounding box of ``person'' and mask part of that bounding box, we can get a prediction of ``person'' but also a new bounding box which contains a different classification. 
We ignore these new bounding boxes and only update the process queue with the bounding box that best matches the prediction and partition which spawned it, as shown in~\Cref{algo:update}.

\begin{algorithm}[hbt]
    \caption{$\mathit{prune}\,(\mathcal{P}, \mathcal{D}, \mathcal{F})$}
    \label{algo:update}
    \begin{flushleft}
        \textbf{INPUT:}\,\, an array of predictions, $\mathit{preds}$, top-level target predictions $\mathcal{D}$, a job queue \emph{Q}, an array $\mathit{updated}$ \\
        \textbf{OUTPUT:}\,\, $\mathit{updated}$, $\mathit{Q}$
    \end{flushleft}     
        \begin{algorithmic}[1]
            \FOR{$\mathit{pred}$ in $\mathit{preds}$}
               \IF{($\mathit{pred} =$ original label in $\mathcal{D})$ \AND $(\mathit{updated(pred)} \neq -1)$}                  
                    \STATE remove all other parts
                    \STATE update location in $\mathit{updated}$
                    \STATE update $\mathit{Q}$ with $\mathit{pred}$
                \ENDIF
            \ENDFOR
            \RETURN $(\mathit{updated}, \mathit{Q})$
        \end{algorithmic}
\end{algorithm}

\begin{figure*}[htb]
    \centering
    \begin{subfigure}[t]{0.23\textwidth}
        \centering
        \includegraphics[scale=0.31]{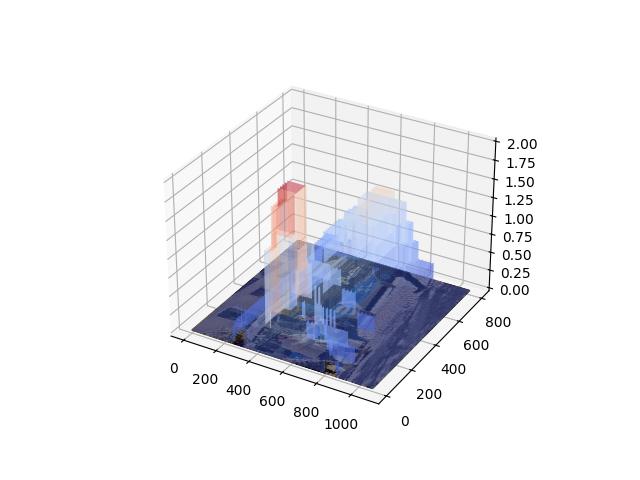}
    \caption{2 iterations}    
    \end{subfigure}
    ~ 
    \begin{subfigure}[t]{0.23\textwidth}
        \centering
        \includegraphics[scale=0.31]{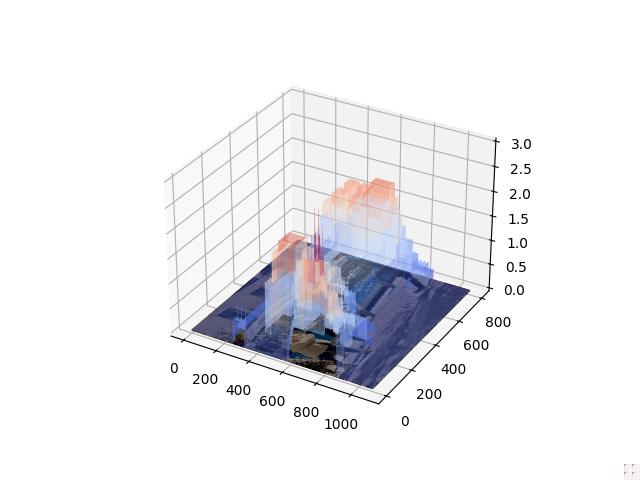}
    \caption{4 iterations}     
    \end{subfigure}
    ~
    \begin{subfigure}[t]{0.23\textwidth}
        \centering
        \includegraphics[scale=0.31]{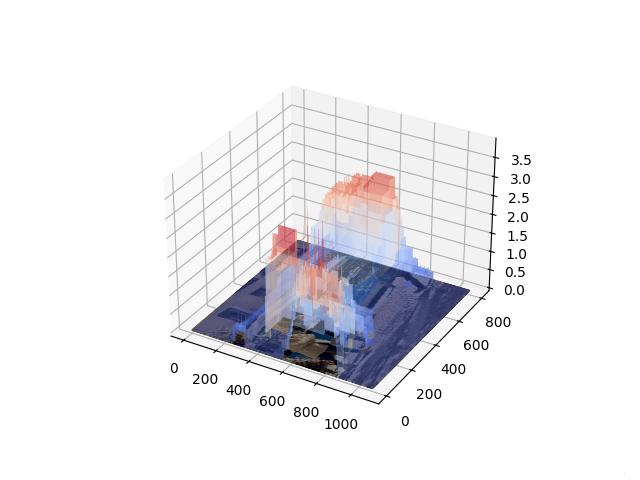}
    \caption{6 iterations}     
    \end{subfigure}    
    ~
    \begin{subfigure}[t]{0.23\textwidth}
        \centering
        \includegraphics[scale=0.31]{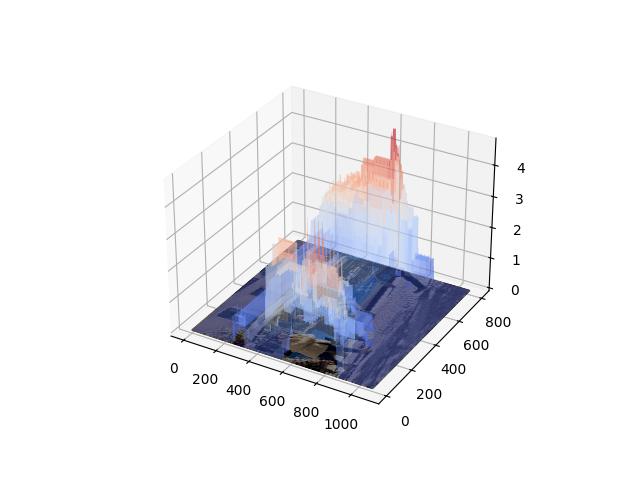}
    \caption{8 iterations}     
    \end{subfigure}  
    \caption{The development of the pixel ranking of the image in~\Cref{fig:workflow} over multiple iterations, using discrete uniform for partition creation. \yolo recognizes $6$ bounding boxes in this image. We build the ranking for all detections together, then separate them when extracting explanations.}\label{fig:blended}
\end{figure*}

After running a set number of iterations to generate the saliency landscape (\Cref{fig:blended}), the explanation extraction procedure 
extracts explanations for all objects by overlaying bounding boxes produced by \yolo on the landscape and adding in pixels, 
grouped by responsibility, under the bounding box, until the explanations satisfy the model, as shown in~\Cref{algo:extract}. 
We sort all unique responsibility values in the bounding box $d$ under investigation in descending order. We then iterate over these different degrees of responsibility. The procedure \emph{add\_pixels\_at()} simply reveals all pixels in the image $x$ which are inside the bounding box for $d$ with the appropriate degree of responsibility. We query the model at each responsibility level under the image passes the classifier. This will always succeed as the original bounding box is a passing classification. We reach the original bounding box when we include all pixels with responsibility $\geq 0$. This situation occurs when the bounding box is already minimal for the \yolo classification.

\begin{algorithm}[hbt]
    \caption{$\mathit{extract}\,(R, \mathcal{D}, \mathcal{N}, $x$)$}
    \label{algo:extract}
    \begin{flushleft}
        \textbf{INPUT:}\,\, a saliency landscape $R$, image target detections $\mathcal{D}$, an object detector $\mathcal{N}$ and an image $x$ \\
        \textbf{OUTPUT:}\,\, an array of explanations $\mathcal{E}$
    \end{flushleft}     
        \begin{algorithmic}[1]
            \STATE $\mathcal{E} \leftarrow \emptyset$
            \FORALL{$d \in D$}
                \STATE $\mathit{mask} \leftarrow \emptyset$
                \STATE $\mathit{levels} \leftarrow \mathit{unique(R, box(d))}$ in descending order
                \FORALL{$\mathit{level} \in \mathit{levels}$}
                    \STATE $\mathit{mask} \leftarrow \mathit{add\_pixels\_at(x, level, box(d))}$\label{line:add}
                    \IF{$\mathcal{N}(\mathit{mask}) = \mathit{class(d)}$}
                        \STATE $\mathcal{E} \leftarrow \mathcal{E} \cup \mathit{mask}$
                    \ENDIF
                \ENDFOR
            \ENDFOR
            \RETURN $\mathcal{E}$
        \end{algorithmic}
\end{algorithm}

In \yorex, extracting explanations from the saliency landscape is complicated not only by presence of bounding boxes, but also by the existence of overlapping bounding boxes. 
If a smaller bounding box overlaps a larger bounding box and has high responsibility, it will tend to be included into the explanation for the larger box, as the larger box dominates the responsibility landscape. This situation leads to noise in explanations for larger bounding boxes, that is, high ranked pixels that actually contribute to a different label. 
We handle this problem by treating the bounding boxes as non-intersecting layers in the saliency landscape, hence ensuring that the explanations are not mixed up. 

\begin{figure}
    \centering
    \includegraphics[scale=1.2]{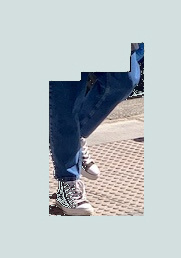}
    \caption{An explanation produced by \yorex for a person in~\Cref{fig:workflow}. The person's entire body is present in the original image.}
    \label{fig:foot}
\end{figure}

\begin{observation}\label{obs}
Explanations found by \yorex might not be the smallest ones.
\end{observation}
It has been observed in the literature that most images contain multiple explanations for their classification~\cite{CKK23,SLKTF21}.
As \yorex finds one explanation per object, which explanations are found depends entirely on the order in which the algorithm computes the degree of responsibility of each part, and hence the output might not be the smallest possible explanation.
We note that this issue does not exist in the original \rex ranking procedure. It arises in \yorex due to the aggressive
pruning, which allows us to significantly reduce the number of queries to the object detector, as
demonstrated in~\Cref{sec:results}. An example for such a non-minimal explanation is shown in 
\Cref{fig:foot}, an explanation for a ``person'' being this person's foot.
While a correct explanation, it seems more likely that a human would choose a face as an explanation rather than a 
foot, given an entire picture of a person. 

%% file: yorex.tex
\tikzset{
   arrowshadow/.style = { thick, color=black, ->, double copy shadow={thick,shadow xshift=2ex, shadow yshift=2ex}},
}

\begin{tikzpicture}[node distance=5.5cm]


\definecolor{tabutter}{rgb}{0.98824, 0.91373, 0.30980}		
\definecolor{ta2butter}{rgb}{0.92941, 0.83137, 0}		
\definecolor{ta3butter}{rgb}{0.76863, 0.62745, 0}		

\definecolor{taorange}{rgb}{0.98824, 0.68627, 0.24314}		
\definecolor{ta2orange}{rgb}{0.96078, 0.47451, 0}		
\definecolor{ta3orange}{rgb}{0.80784, 0.36078, 0}		

\definecolor{tachocolate}{rgb}{0.91373, 0.72549, 0.43137}	
\definecolor{ta2chocolate}{rgb}{0.75686, 0.49020, 0.066667}	
\definecolor{ta3chocolate}{rgb}{0.56078, 0.34902, 0.0078431}	

\definecolor{tachameleon}{rgb}{0.54118, 0.88627, 0.20392}	
\definecolor{ta2chameleon}{rgb}{0.45098, 0.82353, 0.086275}	
\definecolor{ta3chameleon}{rgb}{0.30588, 0.60392, 0.023529}	

\definecolor{taskyblue}{rgb}{0.44706, 0.56078, 0.81176}		
\definecolor{ta2skyblue}{rgb}{0.20392, 0.39608, 0.64314}	
\definecolor{ta3skyblue}{rgb}{0.12549, 0.29020, 0.52941}	

\definecolor{taplum}{rgb}{0.67843, 0.49804, 0.65882}		
\definecolor{ta2plum}{rgb}{0.45882, 0.31373, 0.48235}		
\definecolor{ta3plum}{rgb}{0.36078, 0.20784, 0.4}		

\definecolor{tascarletred}{rgb}{0.93725, 0.16078, 0.16078}	
\definecolor{ta2scarletred}{rgb}{0.8, 0, 0}			
\definecolor{ta3scarletred}{rgb}{0.64314, 0, 0}			

\definecolor{taaluminium}{rgb}{0.93333, 0.93333, 0.92549}	
\definecolor{ta2aluminium}{rgb}{0.82745, 0.84314, 0.81176}	
\definecolor{ta3aluminium}{rgb}{0.72941, 0.74118, 0.71373}	

\definecolor{tagray}{rgb}{0.53333, 0.54118, 0.52157}		
\definecolor{ta2gray}{rgb}{0.33333, 0.34118, 0.32549}		
\definecolor{ta3gray}{rgb}{0.18039, 0.20392, 0.21176}		

\definecolor{pantonegreen}{HTML}{669900}
\definecolor{pantoneblue}{HTML}{0066CC}
\definecolor{pantonered}{HTML}{CC3333}
\definecolor{pantoneorange}{HTML}{FFCC33}

  \node (img) {\includegraphics[width=2cm]{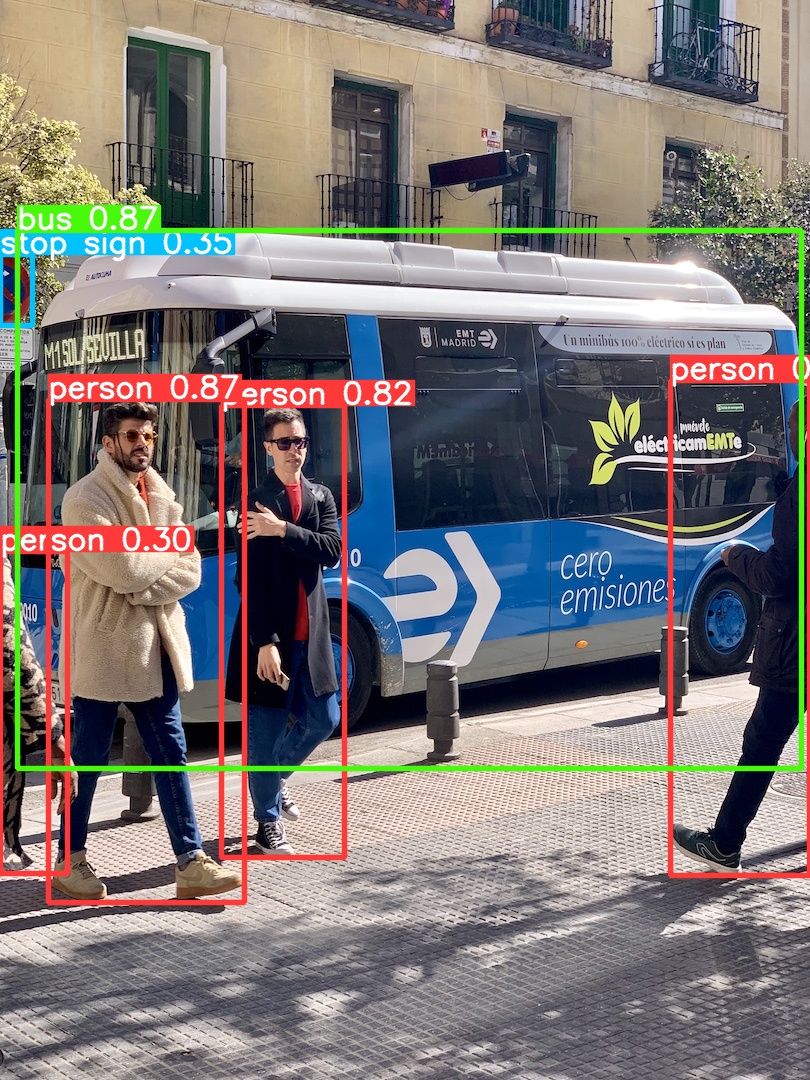}};

  \node[right=0.5cm of img, align=center] (mutants) {\includegraphics[scale=0.18]{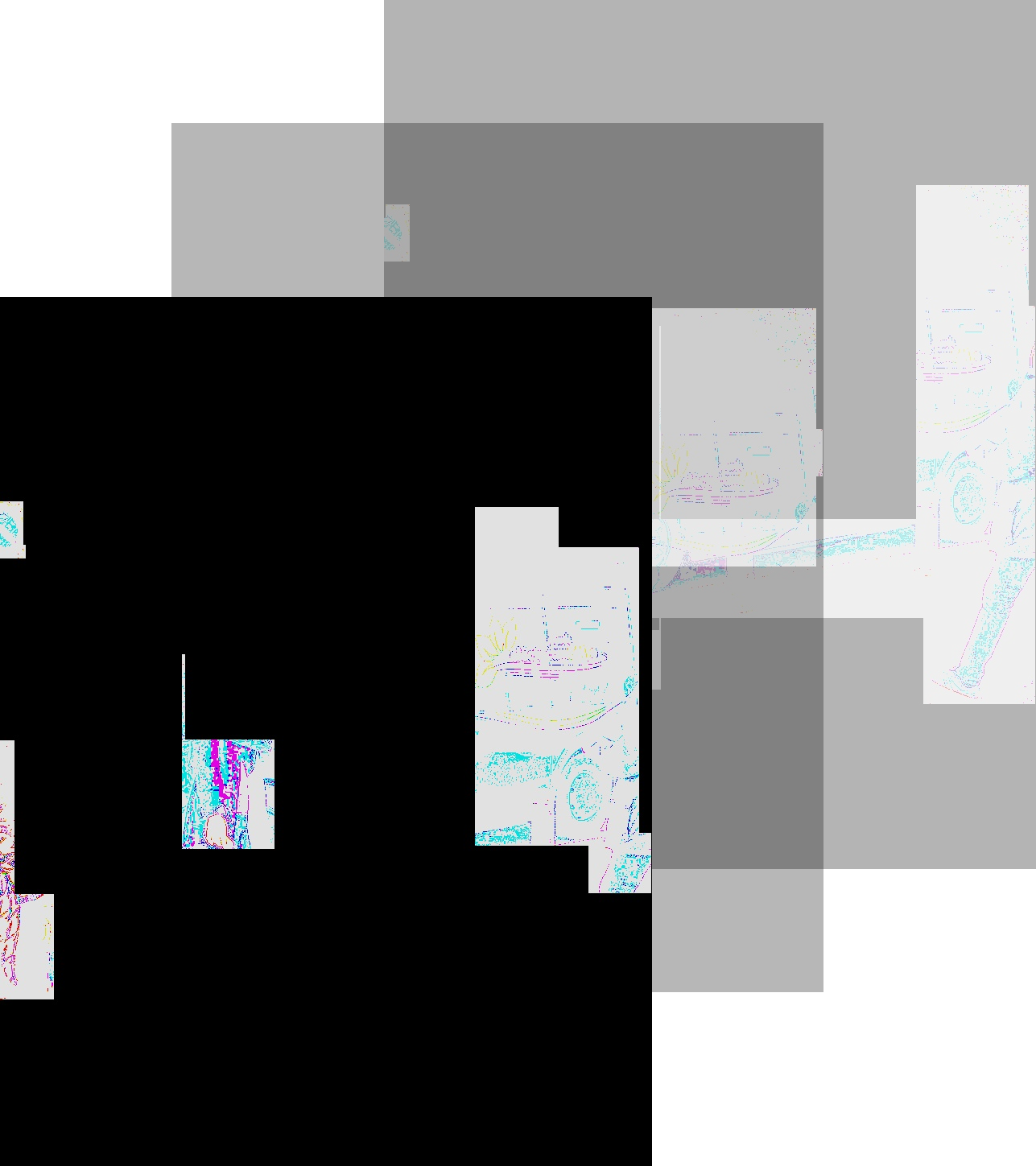}};
  \node[above=0.5mm, align=center] at (mutants.north) {{\large\ding{173}}\\generate\_mutants};

 \node[right=0.5cm of mutants, align=center] (dnn) {\includegraphics[scale=0.08]{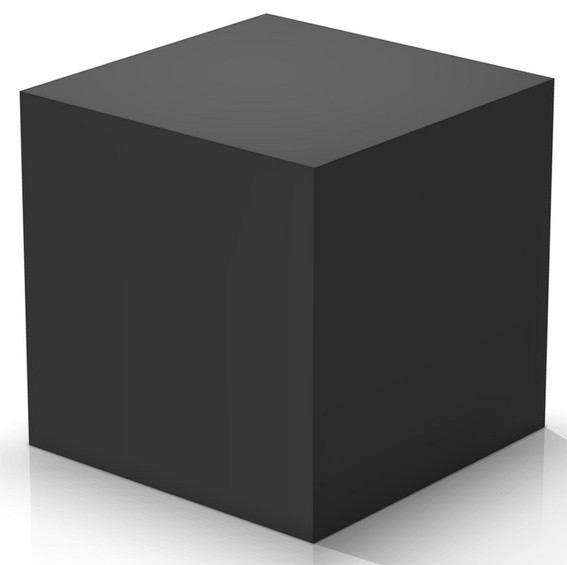}};

  \node[rectangle, rounded corners, top color=tachameleon!60, bottom color=ta2chameleon!50, minimum height=1.75cm, right =0.5cm of dnn, align=center, draw=ta2chameleon] (ra) {{\large\ding{174}}\\Pruning\\Algorithm\\\emph{PRUNE}};

 \draw[->, thick] (img.east)+(-0.2,0.0) -- (1.75, 0.0);

 \draw[->, thick] (mutants.east)+(-0.1,0.0) -- (4.55,0.0); 

 \draw[->, thick] (dnn.east)+(-0.2,0.0) -- (ra);

 \node[right=0.5cm of ra, align=center] (px) {\includegraphics[scale=0.25]{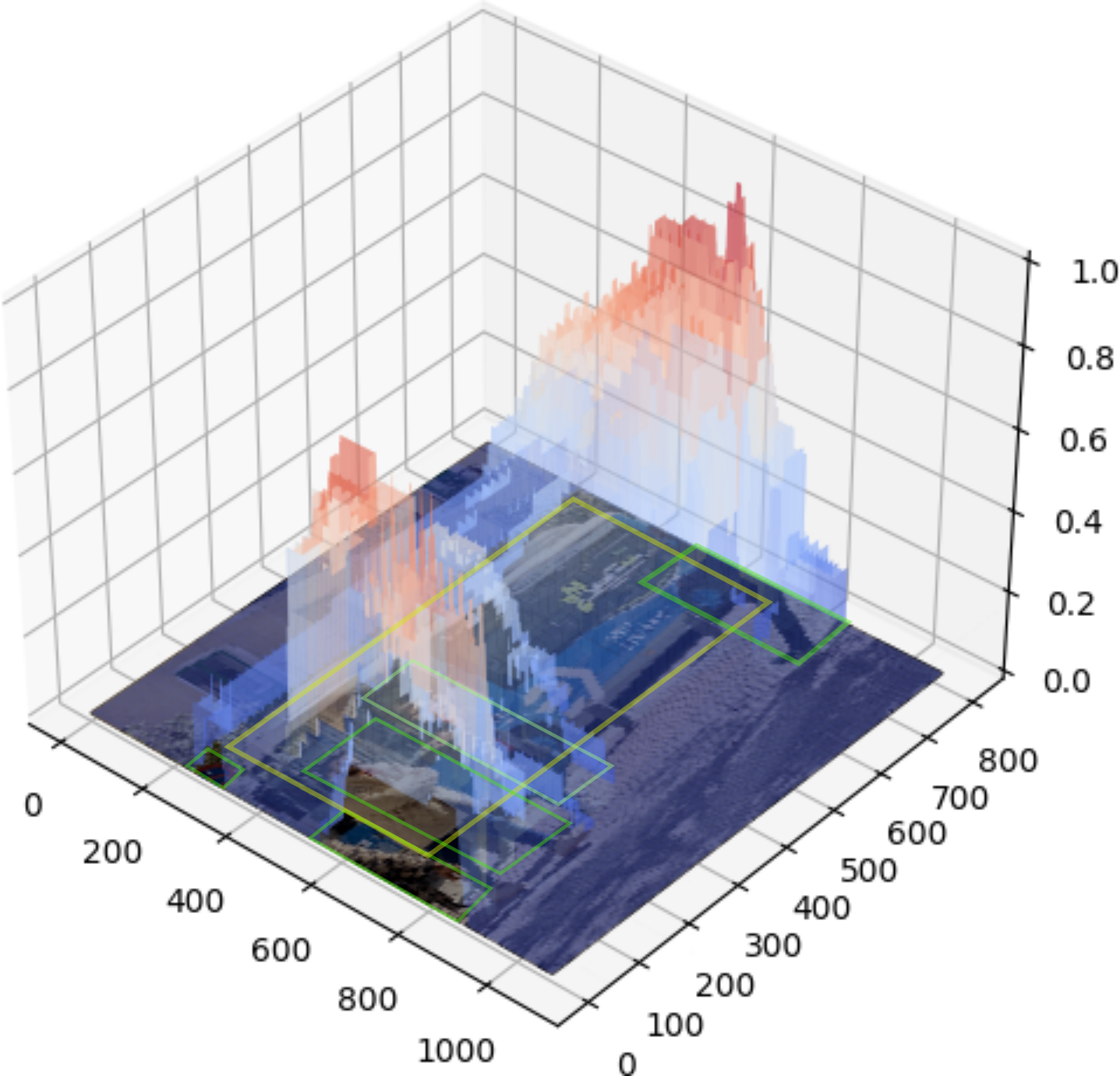}};
 \node[above=-1mm,align=center] at (px.north) {causal\_ranking};

 \draw[->, thick] (ra) -- (9.1,0.0);

  \node[rectangle, rounded corners, align=center, draw=black!50!pantoneorange, fill=pantoneorange!30, minimum height=2cm, right=0.5cm of px] (extract) {\large\ding{175}\\\emph{extract}};

  \draw[->, thick] (px.east)+(-0.1,0.0) -- (extract);

  \draw[->, thick] (px) to [bend left] node [midway, below]{} (mutants);

  \node[rounded corners, align=center, draw=red, fit=(mutants) (px), inner xsep=2mm, inner ysep=8mm,](FIt1){};
  \node[align=center, text height=-3cm] at (FIt1.south){$k$ iterations of inner loop};
 
  \node[inner sep=0pt, right=1cm of extract] (final) {\includegraphics[scale=0.1]{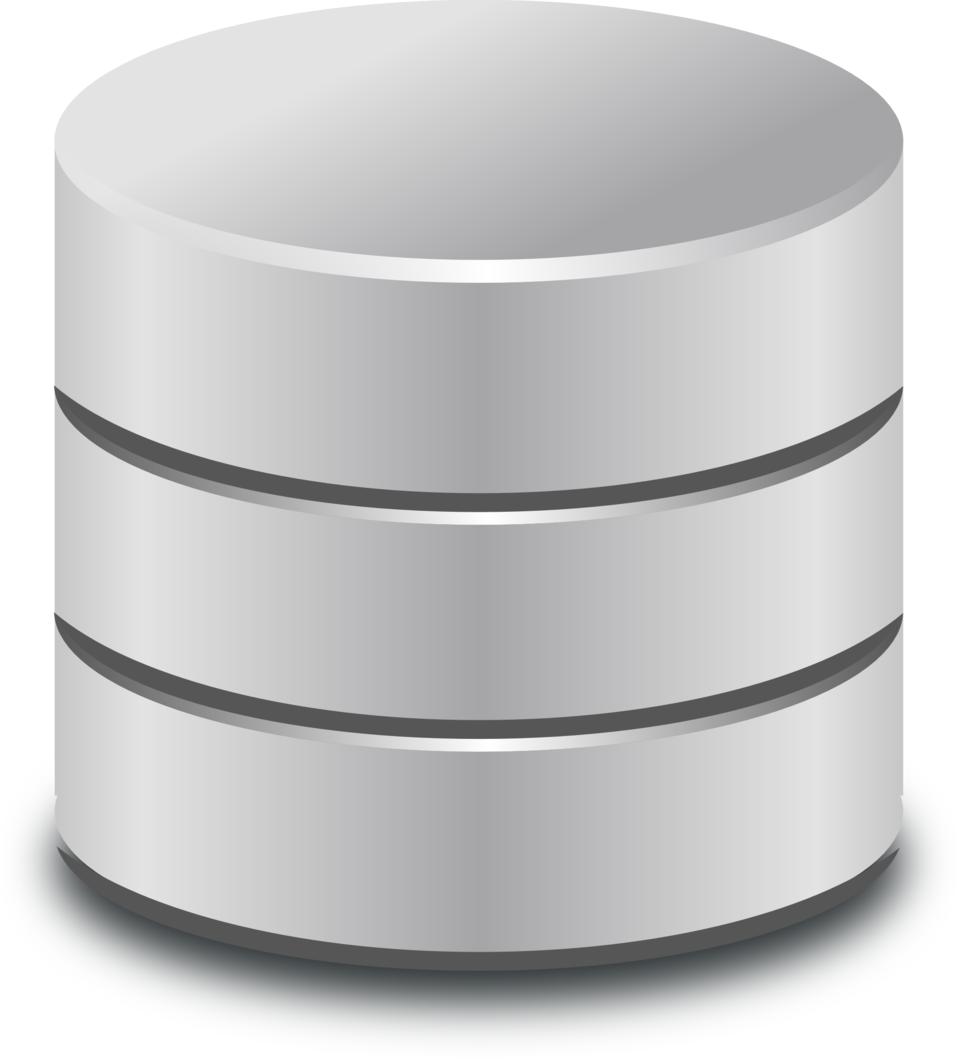}};
   
  \node[rectangle, rounded corners, align=center, right=1.43cm of extract] (final) {\textbf{$\mathcal{E}$}};

  \draw[->, thick] (extract) -- (15.0,0.0);

\end{tikzpicture}

%% file: results.tex
\section{Experiments}\label{sec:results}

\begin{figure*}[htb]
    \centering
    \begin{subfigure}[b]{0.32\textwidth}
        \centering
        \input{yolo_mutants}
    \caption{with \yolo}    
    \end{subfigure}
    \begin{subfigure}[b]{0.32\textwidth}
        \centering
        \input{ssd_mutants}
    \caption{with \ssd}     
    \end{subfigure}
    \begin{subfigure}[b]{0.32\textwidth}
        \centering
        \input{fastrcnn_mutants}
    \caption{with \fastrcnn}     
    \end{subfigure}    
    
    \caption{The average number of calls (proxy for performance) produced by \rex (in red) and our tool \yorex (in blue) using three different 
    object detectors on the ImageNet-mini validation dataset. The performance of our tool is largely independent of the number of objects detected in the image. This in contrast to the performance of \rex, which decreases sharply (that is,
    the number of calls increases) with the increase in the number of objects.}\label{fig:no_mutants}
\end{figure*}
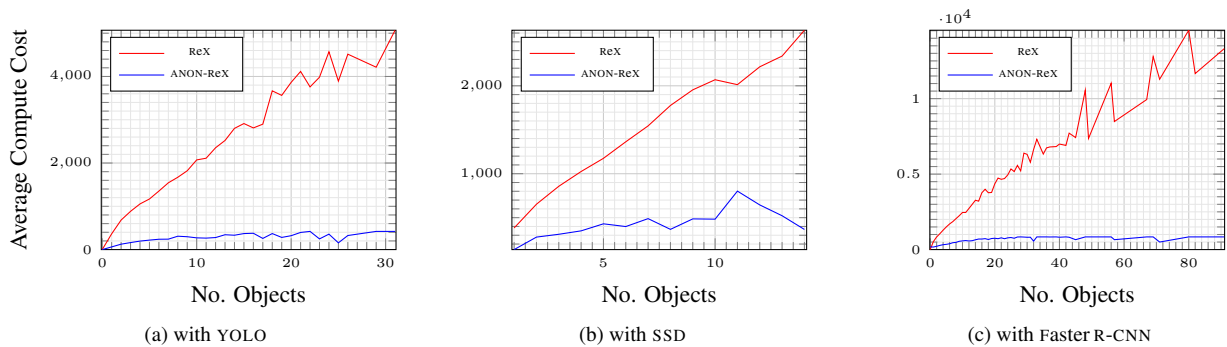

\begin{figure*}[htb]
    \centering
    \begin{subfigure}[t]{0.28\textwidth}
        \centering
        \includegraphics[height=0.6\linewidth, width=1\linewidth]{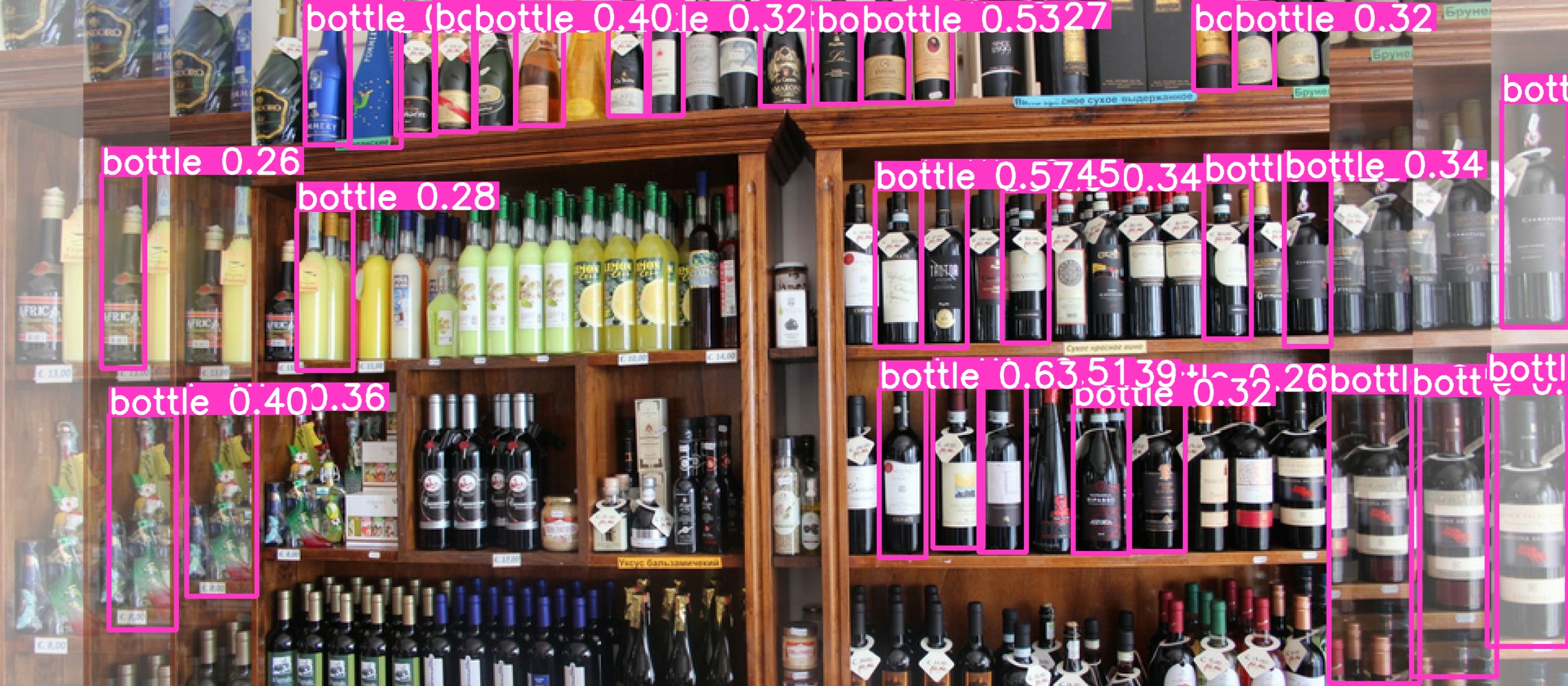}
        \caption{\yolo bounding boxes}\label{subfig:yolowine}
    \end{subfigure}%
    ~ 
    \begin{subfigure}[t]{0.28\textwidth}
        \centering
        \includegraphics[scale=0.13]{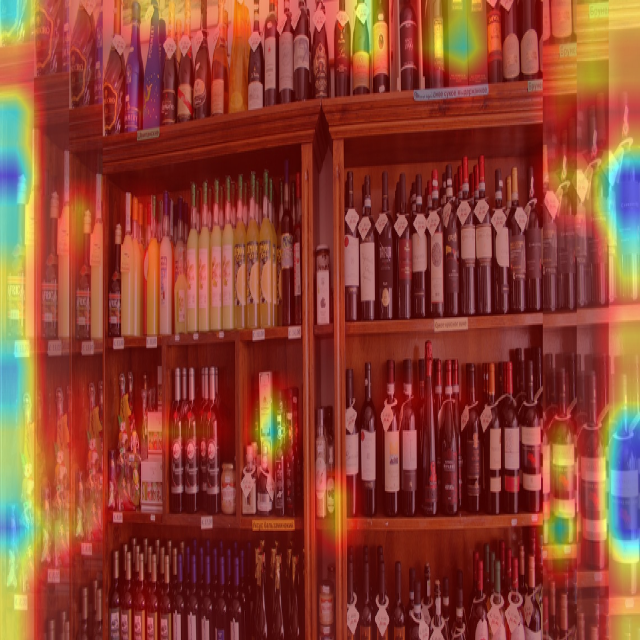}
        \caption{\eigencam{} explanation}\label{subfig:eigencamwine}
    \end{subfigure}
    ~
    \begin{subfigure}[t]{0.28\textwidth}
        \centering
        \includegraphics[height=0.6\linewidth, width=1\linewidth]{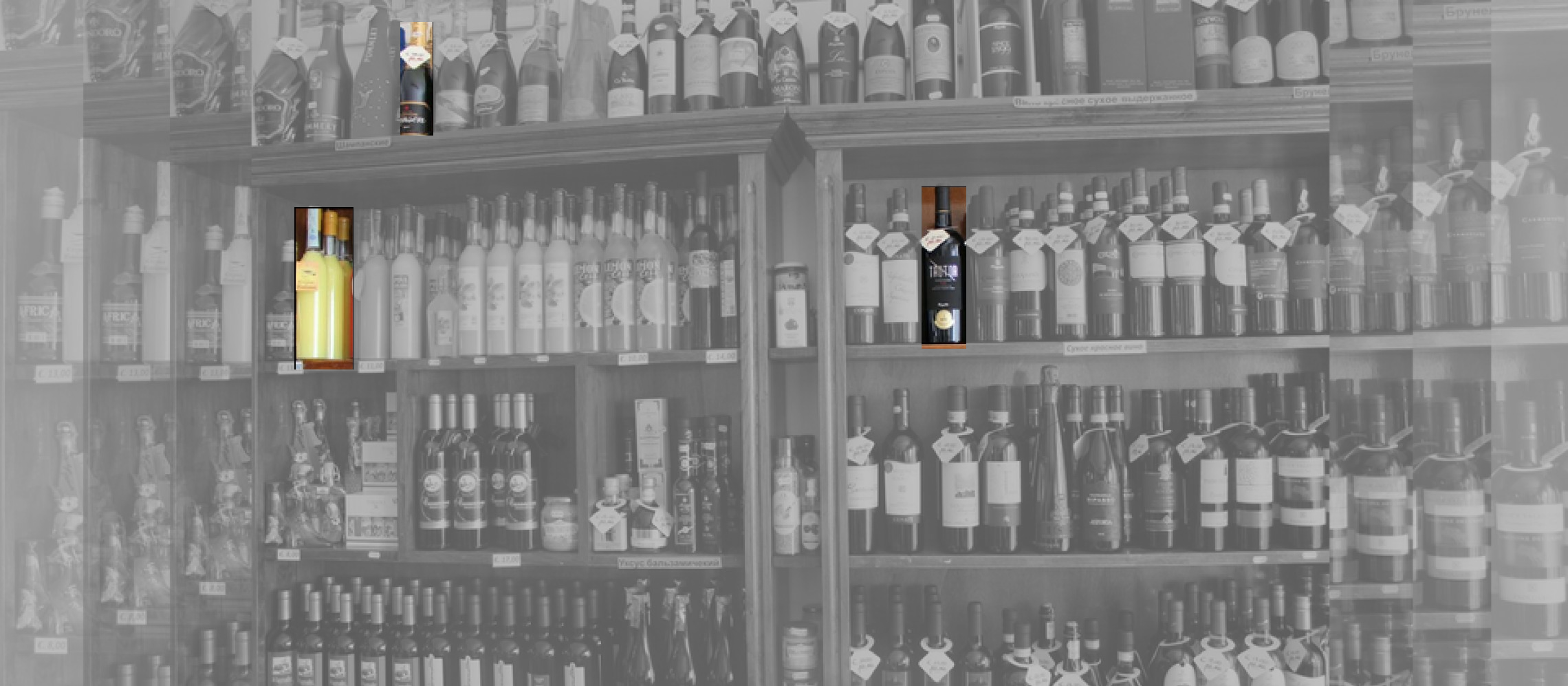}
        \caption{Some of the \yorex explanations}\label{subfig:yorexwine}
    \end{subfigure}    
    \caption{The \yolo bounding boxes~\ref{subfig:yolowine} and explanations as provided by \eigencam{}~\ref{subfig:eigencamwine}. Due to the large number of bottles, the resulting explanation is almost entirely red. While technically correct, this is not very useful. \ref{subfig:yorexwine} show a selection of \yorex explanations.~\Cref{subfig:eigencamwine} has been resized as per the tool's documentation.}\label{fig:wine}
\end{figure*}


In this section we present our experimental results for \yorex. We compare \yorex with \rex on the same object detectors, using the number of queries (number of mutant images) as a proxy
for performance. Both tools are evaluated on \yolo, \ssd, and \fastrcnn.
\begin{table}[htb]
    \centering
    \begin{tabularx}{1\linewidth}{m{0.15\linewidth}||r|r|r|r|r|r}
         & \multicolumn{6}{c}{Number of Objects} \\
        \cmidrule{2-7}
          & \multicolumn{2}{c|}{\yolo} & \multicolumn{2}{c|}{\ssd} & \multicolumn{2}{c}{\textsc{f}\textnormal{aster}} \\
          & \multicolumn{2}{c|}{} & \multicolumn{2}{c|}{} & \multicolumn{2}{c}{\textsc{r}-\textsc{cnn}} \\
          \cmidrule{2-7}\cmidrule{2-7}
           & $10$ & $20$ & $10$ & $13$ & $10$ & $20$ \\
           \midrule
       baseline  & 2071 & 3864 & 2069 & 2338 & 530 & 4358 \\
       \yorex  & 271    & 321  & 483  & 522 & 170 & 751 \\
    \end{tabularx}
    \caption{The number of queries (rounded to the nearest integer) produced by the baseline and our tool \yorex for explaining object detector output, using queries as a proxy for performance.
    We present the results for $10$ and for $20$ objects, except for \ssd, which only recognizes up to $13$ objects. Smaller numbers are better.}
    \label{tab:boxes}
\end{table}

For all experiments
we use the pre-trained yolov8n model~\cite{yolov8_ultralytics} on the standard validation dataset 
ImageNet-mini~\cite{imagenetmini}, consisting of 
$3923$ images. If \yolo v8n cannot perform any detection in an image, we exclude that image from consideration. In our experiments, this occurred only for $210$ images.

\subsection{Analysis of \eigencamyolo}
A direct comparison with \eigencamyolo is challenging, as \eigencamyolo only produces heatmaps. 
Beyond needing post-processing to extract explanations, a heatmap is a not very useful 
when there is a very large number of bounding boxes in an image. \Cref{subfig:yolowine} shows a \yolo prediction 
for an image containing many bottles. The resulting heatmap produced by \eigencamyolo is in~\Cref{subfig:eigencamwine}. Clearly, having almost the whole image marked as ``hot'' does not provide a lot of insight into the \yolo algorithm.

\begin{figure}[htb]
    \centering
    \begin{subfigure}[t]{0.2\textwidth}
        \centering
        \includegraphics[scale=0.15]{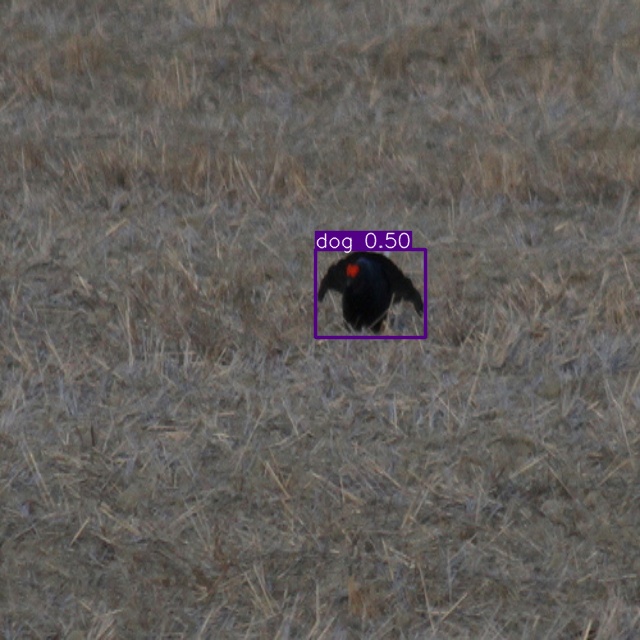}
        \caption{\yolo bounding box for a dog.}\label{subfig:yolodog}
    \end{subfigure}%
    ~ 
    \begin{subfigure}[t]{0.2\textwidth}
        \centering
        \includegraphics[scale=0.15]{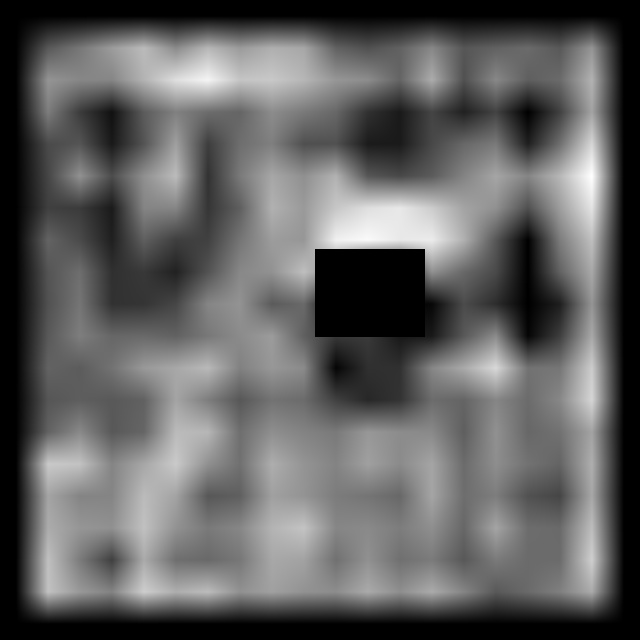}
        \caption{The greyscale heatmap generated by \eigencamyolo outside of \yolo's bounding box.}\label{subfig:eigendog}
    \end{subfigure}
    \caption{The large proportion of non-black pixels outside the bounding box in~\ref{subfig:eigendog} shows that \eigencamyolo's explanation takes up the entire image frame, 
    rather than fitting inside the bounding box described by \yolo.}\label{fig:eigendog}
\end{figure}

Our quantitative analysis of \eigencamyolo is based on a non-controversial assumption
that a good explanation should reside inside the corresponding bounding box, and be at maximum the size of the bounding box. This is because the bounding box 
contains all the information necessary for the classification, in addition to some extraneous pixels due to the rectangular shape of bounding boxes. In order to measure the efficiency of \eigencamyolo's explanations, we count the number of ``hot'' pixels detected by \eigencamyolo that reside outside of the bounding boxes determined by \yolo, and normalize the count by the total number of pixels in the image. A normalized score of $0$ indicates no superfluous pixels, while a normalized score closer to $1$ indicates that almost all of the ``hot'' pixels fall outside of \yolo's bounding boxes. An example of these superfluous pixels is shown in~\Cref{fig:eigendog}. This particular image was one of the worst detected. We used the validation set from ImageNet-mini for this analysis.

On average, the percentage of ``hot'' pixels detected in an image by \eigencamyolo that fall outside of the \yolo bounding boxes is $30.5\%$. 
In other words, about $30\%$ of the explanation created by \eigencamyolo for a particular image exceeds the target maximal explanation produced by \yolo.
On the basis of this excessive size, we excluded \eigencamyolo from the comparative analysis.

\subsection{Experimental results}

We use \rex as a baseline black-box explainability tool to which we compare the performance of \yorex. Recall that even black-box tools require a non-trivial adaptation to the
output format of object detectors, and to the best of our knowledge, no such adaptations exist. 
In order to use basic \rex with object detectors, we introduced changes to the input format of \rex without modifying the algorithm. 
Specifically, for each bounding box detected, we produced a new image in which everything but the bounding box is set to the masking color. We then executed \rex over each independent image (\ie, $6$ bounding boxes result in $6$ independent runs of \rex). We also added code to process the results of \yolo and \ssd, as they are in a different format
than the one expected by \rex. Finally, \yolo can produce a ``no classification'' output, which \rex does not expect. 
If \yolo is unable to identify anything in the image, we ignore it and do not call \rex on these outputs.

We use the number of mutants generated by both tools as a proxy for performance. As the internal algorithm of \yorex takes a negligible amount of time,
this is a reasonable approximation. In fact, \yorex is faster than \rex even on the same number of mutants, 
as it also makes much more aggressive use of batching than the original \rex implementation.

\Cref{fig:no_mutants} shows the average number of mutants produced as a function of the number of objects detected in the image for the three object detectors. 
Beyond demonstrating the superior performance of \yorex, it also
shows that \yorex performance is not affected by the number of objects detected in the image, in contrast to \rex, whose performance rapidly deteriorates
with the increase in the number of objects.

A sample of the experimental results is also presented in~\Cref{tab:boxes}, with the performance for images with $10$ objects and for $20$ objects (except for \ssd, where the maximal number of objects
is $13$), compared to the baseline (\rex). The results show at least an order of magnitude speedup, with \yorex scaling to multiple objects with only a minimal increase in the number of queries. 

\begin{remark}
The size of produced explanations is often used as a proxy for quality. In our experiments, explanations produces by \yorex are $20.7\%$ larger than those produced by \rex for \yolo,
$22.4\%$ larger for \ssd, and $52.9\%$ larger for \fastrcnn. This is partly due to the aggressive pruning, and partly due to object detectors not recognizing small fragments of images as well as image
classifiers.
\end{remark}

\subsection{Explaining images with many objects}\label{subsec:many}

Recall that~\Cref{subfig:yolowine} shows a \yolo prediction for an image containing many bottles. 
\yorex explanations are provided in the form of subsets of pixels, where each subset is sufficient in each bounding box to 
generate the same label, though not with the same confidence. For the image of many bottles, shown in \Cref{subfig:yolowine} with
the \yolo predictions, ~\Cref{subfig:yorexwine} shows the explanations generated by \yorex. Of particular interest is the yellow bottle explanation on the left. Closer inspection reveals that there are several bottles in this explanation, not all of them yellow; 
indeed, on the other yellow bottles in the image, \yolo fails to classify them at all. 
This may indicate that \yolo is also relying partially on color and not just shape to distinguish a bottle. 
\begin{figure}
    \centering
    \includegraphics[scale=0.5]{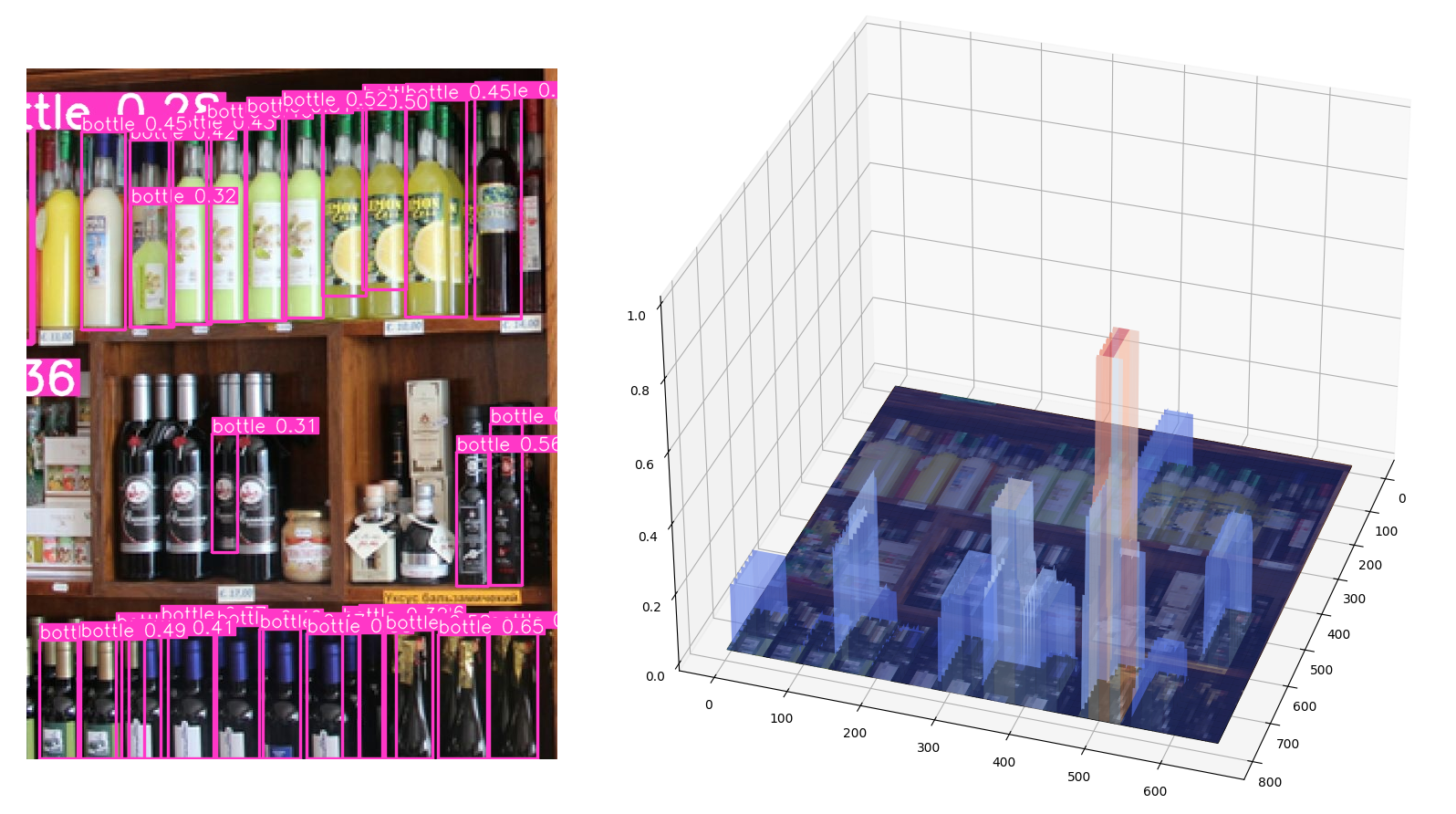}
    \caption{The \yorex pixel ranking for the cropped version of the bottles in~\Cref{fig:wine}. \yolo discovers new bottles in the image, but the saliency landscape for the yellow bottles is flat, indicating that no subdivision of the bounding box was detected as a bottle. This in contrast with the colored bottles on the bottom row, all of which have explanations smaller than their original bounding boxes.}
    \label{fig:bottle_fail}
\end{figure}
We cropped the image to include only the section in which \yolo did not identify any objects and re-executed \yolo and \yorex on it. 
\Cref{fig:bottle_fail} shows that \yolo finds the yellow bottles on the cropped image. However those have a low confidence, and the
saliency landscape for this area is almost completely flat, indicating that there is no explanation for any of these bottles that is smaller than
the bounding box itself. This contrasts with the partial, dark, bottles on the bottom row, which have been successfully subdivided at 
least once, even though they are only partially visible.
This indicates \yolo's dependency on color which, to a human, is orthogonal to recognizing a bottle.

\commentout{
\subsection{Non-obvious explanations}\label{subsec:nonobvious}
As mentioned in~\Cref{obs}, \yorex might miss a smaller explanation in favor of a larger one. Furthermore, these explanations might not be the most obvious ones, in particular for objects
that have a number of explanations for their label.
\Cref{fig:foot} shows one such example, of an explanation for a ``person'' being this person's foot.
While a correct explanation, it seems more likely that a human would choose a face as an explanation rather than a foot, given an entire picture of a person. 

\begin{figure}
    \centering
    \includegraphics[scale=1.8]{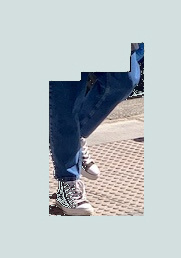}
    \caption{A sufficient, but not necessary, explanation for a person from \yorex, as found in the image in~\Cref{fig:workflow}. The person's entire body is present in the original image.}
    \label{fig:foot}
\end{figure}
} 

\subsection{Shape of explanations}\label{subsec:round}

All object detectors we examined produce rectangular bounding boxes. In our experiments, the explanations provided by \yorex were rectilinear, even when the objects
do not have straight lines and edges. This may be due to the requirements of \yolo, an insufficient number of iterations from \yorex, or a combination of the two. To investigate this issue, we executed \yorex, with \yolo as the object detector, on a dataset containing images of apples~\cite{apples}. Apples were
chosen as objects without straight lines and ones where the ground truth is known. \Cref{fig:applepx} in the introduction shows an example of some of its outputs. 
Inspection of the results indicates that even with $20$ iterations, the shape of \yorex explanations is strongly influenced by the bounding box requirements. 
An explanation must be confirmed by the model, and \yolo and similar tools require enough pixels to draw a rectangle, a fundamental part of their classification,
thus limiting the shape of the final explanation. On the other hand, \yorex explanations can be much smaller than \yolo bounding boxes, hence providing insights
into the \yolo detection process.

\begin{figure}[htb]
    \centering
    \begin{subfigure}[t]{0.2\textwidth}
        \centering
        \includegraphics[scale=0.18]{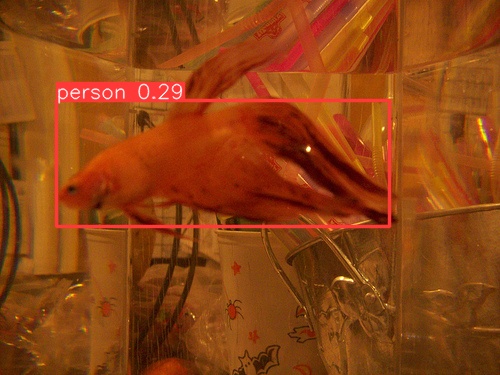}
        \caption{\yolo classification}\label{subfig:misfish}
    \end{subfigure}%
    ~ %
    \begin{subfigure}[t]{0.2\textwidth}
        \centering
        \includegraphics[scale=0.75]{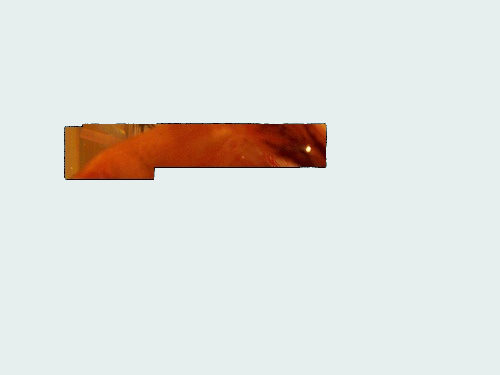}
        \caption{\yorex explanation}\label{subfig:exmisfish}
    \end{subfigure}
    \caption{\ref{subfig:exmisfish} suggests that \yolo misclassifies a fish as a human in \ref{subfig:misfish},
because of the part of the fish that is mistaken for red hair.}\label{fig:misfish}
\end{figure}

\begin{figure}[htb]
    \centering
    \begin{subfigure}[t]{0.18\textwidth}
        \centering
        \includegraphics[scale=0.18]{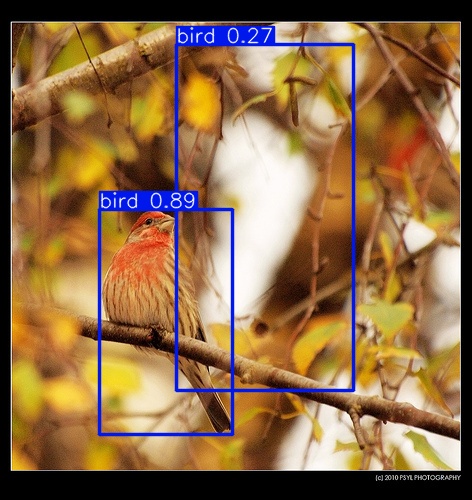}
        \caption{\yolo bird classification}\label{subfig:yolobird}
    \end{subfigure}    
       ~
    \begin{subfigure}[t]{0.18\textwidth}
        \centering
        \includegraphics[scale=0.75]{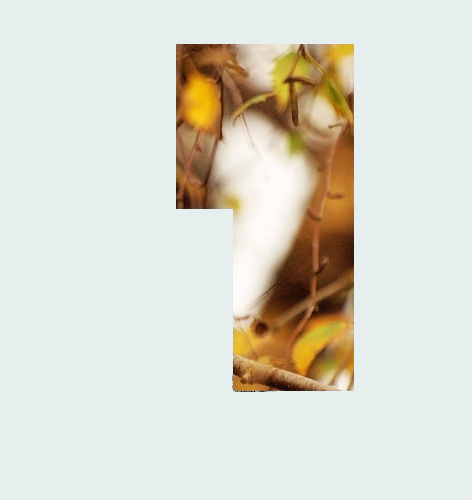}
        \caption{\yorex shows a misclassification}\label{subfig:misbird}
    \end{subfigure} 
    \caption{The image in \ref{subfig:yolobird} is misclassified as two birds, with the explanation in \ref{subfig:misbird} showing that
a branch and white patch in the image are being interpreted as a bird.}\label{fig:misbird}
\end{figure}

\subsection{Explaining misclassifications}\label{subsec:misclassifications}
\yolo, like all models, is not $100\%$ accurate. We analyse two examples of misclassifications and show that their explanations, produced by \yorex, can be helpful in identifying the
reasons for misclassification. 
\yolo recognizes $80$ different classes and is capable of returning ``no classification'' if it does not recognize anything in the image. \Cref{fig:misfish} shows an image of an object that \yolo does not know: a fish. It still provides a very good bounding box detection on the image, with a low confidence, $0.27$, that the fish is a person. The \yorex explanation (\Cref{subfig:exmisfish}) highlights the upper region of the fish, suggesting that \yolo is interpreting the image as containing (red) hair. \Cref{fig:misbird} shows an object from a class that \yolo does understand: a bird. Again, we see a bounding box with low confidence labelled bird. The explanation from \yorex is apparently irreducible, being the same size as the original bounding box. \yolo has found something that is plausibly a bird, as the explanation shows both a branch at the bottom of the image (for perching) and a large white shape that swells and contracts much in the fashion of a bird's body. In our datasets we were unable to identify a misclassification which also had a high confidence.

%% file: yolo_mutants.tex
\begin{tikzpicture}
    \begin{axis}[
    grid=both,
    height=4.5cm,
    width=5.5cm,
    minor tick num=9,
    grid style={line width=.2pt, draw=black!10},
         major grid style={line width=.4pt,draw=black!20},
    legend style={font=\tiny},   
     enlargelimits={abs=0.1},
	   ylabel=\small{Average Compute Cost},
	   xlabel=\small{No. Objects},
legend pos=north west,
        ]
    \addplot[color=red] coordinates {
(0,     0) 
(1,   364) 
(2,   682) 
(3,   884) 
(4,  1055) 
(5,  1169) 
(6,  1352) 
(7,  1543) 
(8,  1671) 
(9,  1821) 
(10,  2072)
(11,  2111)
(12,  2353)
(13,  2521)
(14,  2801)
(15,  2912)
(16,  2811)
(17,  2898)
(18,  3668)
(19,  3563)
(20,  3864)
(21,  4116)
(22,  3759)
(23,  3983)
(24,  4571)
(25,  3892)
(26,  4515)
(29,  4214)
(31,  5068)
    };

\addplot[color=blue] coordinates {
(0,    0)
(1,   62)
(2,  125)
(3,  162)
(4,  196)
(5,  219)
(6,  239)
(7,  241)
(8,  309)
(9,  297)
(10,  271)
(11,  265)
(12,  278)
(13,  343)
(14,  332)
(15,  370)
(16,  377)
(17,  258)
(18,  370)
(19,  281)
(20,  321)
(21,  396)
(22,  420)
(23,  247)
(24,  357)
(25,  156)
(26,  325)
(29,  420)
(31,  420)
};
\legend{\rex, \yorex}
    \end{axis}
\end{tikzpicture}

%% file: ssd_mutants.tex
\begin{tikzpicture}
    \begin{axis}[
    grid=both,
    height=4.5cm,
    width=5.5cm,
    minor tick num=9,
    grid style={line width=.2pt, draw=black!10},
         major grid style={line width=.4pt,draw=black!20},
    legend style={font=\tiny},   
     enlargelimits={abs=0.1},
	   xlabel=\small{No. Objects},
legend pos=north west,
        ]
    \addplot[color=red] coordinates {
(0   0)   
(1, 383)  
(2, 653)  
(3, 857)  
(4, 1025) 
(5, 1175) 
(6, 1363) 
(7, 1544) 
(8, 1776) 
(9, 1954) 
(10, 2069)
(11, 2013)
(12, 2216)
(13, 2338)
(14, 2632)
    };

\addplot[color=blue] coordinates {
(0  0)    
(1, 137)
(2, 279)
(3, 311)
(4, 350)
(5, 430)
(6, 399)
(7, 489)
(8, 367)
(9, 486)
(10, 483)
(11, 802)
(12, 645)
(13, 522)
(14, 366)
};
\legend{\rex, \yorex}
    \end{axis}
\end{tikzpicture}

%% file: fastrcnn_mutants.tex
\begin{tikzpicture}
    \begin{axis}[
    grid=both,
    height=4.5cm,
    width=5.5cm,
    minor tick num=9,
    grid style={line width=.2pt, draw=black!10},
         major grid style={line width=.4pt,draw=black!20},
    legend style={font=\tiny},   
     enlargelimits={abs=0.1},
	   xlabel=\small{No. Objects},
legend pos=north west,
        ]
    \addplot[color=red] coordinates {
 (0,      0)
 (1,    529)
 (2,    836)
 (3,   1052)
 (4,   1285)
 (5,   1502)
 (6,   1697)
 (7,   1853)
 (8,   2051)
 (9,   2244)
(10,   2459)
(11,   2459)
(12,   2725)
(13,   2995)
(14,   3269)
(15,   3207)
(16,   3767)
(17,   3995)
(18,   3767)
(19,   3783)
(20,   4358)
(21,   4738)
(22,   4656)
(23,   4705)
(24,   4944)
(25,   5342)
(26,   5171)
(27,   5579)
(28,   5216)
(29,   6391)
(30,   6309)
(31,   5786)
(32,   6627)
(33,   7308)
(35,   6328)
(36,   6737)
(37,   6799)
(39,   6825)
(40,   6991)
(42,   6902)
(43,   7714)
(45,   7420)
(48,  10556)
(49,   7364)
(56,  11018)
(57,   8484)
(67,   9940)
(69,  12782)
(71,  11284)
(80,  14532)
(82,  11662)
(91,  13314)
    };

\addplot[color=blue] coordinates {
 (0,   180)
 (1,   169)
 (2,   218)
 (3,   279)
 (4,   324)
 (5,   348)
 (6,   386)
 (7,   455)
 (8,   479)
 (9,   550)
(10,   576)
(11,   599)
(12,   570)
(13,   589)
(14,   650)
(15,   699)
(16,   698)
(17,   725)
(18,   669)
(19,   733)
(20,   751)
(21,   722)
(22,   779)
(23,   720)
(24,   778)
(25,   800)
(26,   751)
(27,   834)
(28,   840)
(29,   830)
(30,   816)
(31,   819)
(32,   558)
(33,   840)
(35,   840)
(36,   840)
(37,   835)
(39,   840)
(40,   818)
(42,   840)
(43,   818)
(45,   656)
(48,   840)
(49,   840)
(56,   840)
(57,   656)
(67,   840)
(69,   840)
(71,   500)
(80,   840)
(82,   840)
(91,   840)
};
\legend{\rex, \yorex}
    \end{axis}
\end{tikzpicture}

%% file: conclusions.tex
\section{Conclusions}\label{sec:conclusion}
\yorex is the first black box XAI tool for explaining the outputs of object detectors. It significantly outperforms \rex, so the impact of \yorex explanations on 
the performance of object detectors is much lower, opening the possibility for using \yorex for explainability in real time. 
We demonstrated the generality of our method and its efficiency by using \yorex on several different object detection tools, both one-pass and two-pass, namely, \yolo, \ssd, and \fastrcnn.
We show that existing white box explainability methods, in addition to requiring a significant amount of coding, do not produce sufficiently precise results in terms of quality of explanations.